\documentclass[lettersize,journal]{IEEEtran}
\usepackage{amsmath,amsfonts}
\usepackage{algorithmic}
\usepackage{array}
\usepackage{textcomp}
\usepackage{stfloats}
\usepackage{url}
\usepackage{verbatim}
\usepackage{graphicx}
\usepackage{cite}
\usepackage{booktabs}
\usepackage{amssymb}
\usepackage{bbding}
\usepackage{pifont}
\usepackage{wasysym}
\usepackage{utfsym}
\usepackage{fontawesome}
\usepackage[inline]{enumitem}
\usepackage{colortbl}
\usepackage{bm}
\usepackage{multirow}
\usepackage{subcaption}
\usepackage{graphicx}

\usepackage[normalem]{ulem}
\usepackage[linesnumbered,ruled,vlined]{algorithm2e}
\usepackage{xcolor}
\definecolor{vgray}{gray}{0.88}

\begin{document}

\title{Learning from Semi-Factuals: A Debiased and Semantic-Aware Framework for Generalized Relation Discovery }

\author{Jiaxin Wang, Lingling Zhang, Jun Liu, Tianlin Guo, Wenjun Wu

\thanks{ Jiaxin Wang, Lingling Zhang, Jun Liu are with the School of Computer Science and Technology, Xi'an Jiaotong University, Xi'an, Shaanxi 710049, China (e-mail: jiaxinwangg@outlook.com; zhanglling@xjtu.edu.cn; liukeen@xjtu.edu.cn).}

\thanks{Tianlin Guo and Wenjun Wu are with the Shaanxi Provincial Key Laboratory of Big Data Knowledge Engineering, Xi'an Jiaotong University, Xi'an, Shaanxi 710049, China and also with National Engineering Lab for Big Data Analytics, Xi'an Jiaotong University, Xi'an, Shaanxi 710049, China (e-mail: 1072446164@stu.xjtu.edu.cn; nickjunwork@163.com).}

\thanks{This work has been submitted to the IEEE for possible publication. Copyright may be transferred without notice, after which this version may no longer be accessible.}}

\markboth{Journal of \LaTeX\ Class Files,~Vol.~, No.~}%
{Shell \MakeLowercase{\textit{et al.}}: A Sample Article Using IEEEtran.cls for IEEE Journals}



\maketitle

\begin{abstract}
We introduce a novel task, called Generalized Relation Discovery (GRD), for open-world relation extraction. GRD aims to identify unlabeled instances in existing pre-defined relations or discover novel relations by assigning instances to clusters as well as providing specific meanings for these clusters. The key challenges of GRD are how to mitigate the serious model biases caused by labeled pre-defined relations to learn effective relational representations and how to determine the specific semantics of novel relations during classifying or clustering unlabeled instances. We then propose a novel framework, SFGRD, for this task to solve the above issues by learning from semi-factuals in two stages. The first stage is semi-factual generation implemented by a tri-view debiased relation representation module, in which we take each original sentence as the main view and design two debiased views to generate semi-factual examples for this sentence. The second stage is semi-factual thinking executed by a dual-space tri-view collaborative relation learning module, where we design a cluster-semantic space and a class-index space to learn relational semantics and relation label indices, respectively. In addition, we devise alignment and selection strategies to integrate two spaces and establish a self-supervised learning loop for unlabeled data by doing semi-factual thinking across three views. Extensive experimental results show that SFGRD surpasses state-of-the-art models in terms of accuracy by 2.36\% $\sim$5.78\% and cosine similarity by 32.19\%$\sim$ 84.45\% for relation label index and relation semantic quality, respectively. To the best of our knowledge, we are the first to exploit the efficacy of semi-factuals in relation extraction.
\end{abstract}


\section{Introduction}
\label{introduction}
Relation Extraction (RE) is an essential technology for knowledge graph construction that identifies the relation between two entities within a sentence~\cite{DBLP:journals/tkde/LiangLLZSG23}. For example, when given the entity pair \emph{(New Orleans, the United States)} in the sentence \emph{New Orleans is located in the United States}, a relation extraction model should predict the relation \emph{located in} between these two entities. Although conventional RE models have made significant progress~\cite{DBLP:conf/acl/LiuLHC022, DBLP:journals/tkde/QuHOZ23}, they are designed based on pre-defined relations and cannot handle novel relations in the open world. Open Relation Extraction (OpenRE) has emerged with the goal of extracting relations without pre-defined types or annotated data in the open-world corpus. To achieve this, OpenRE is commonly formulated as an unsupervised clustering task~\cite{DBLP:conf/acl/TranLA20, DBLP:conf/emnlp/HuWXZY20,DBLP:conf/emnlp/WuYHXLLLS19,DBLP:conf/emnlp/ZhaoGZZ21,DBLP:conf/emnlp/LiJH22, DBLP:conf/acl/ZhaoZZGWPS23}, which clusters relational instances into groups with their similarity in a metric space and regards each cluster as a relation. 
\begin{figure}[t]
	\centering
	\includegraphics[width=0.9\linewidth]{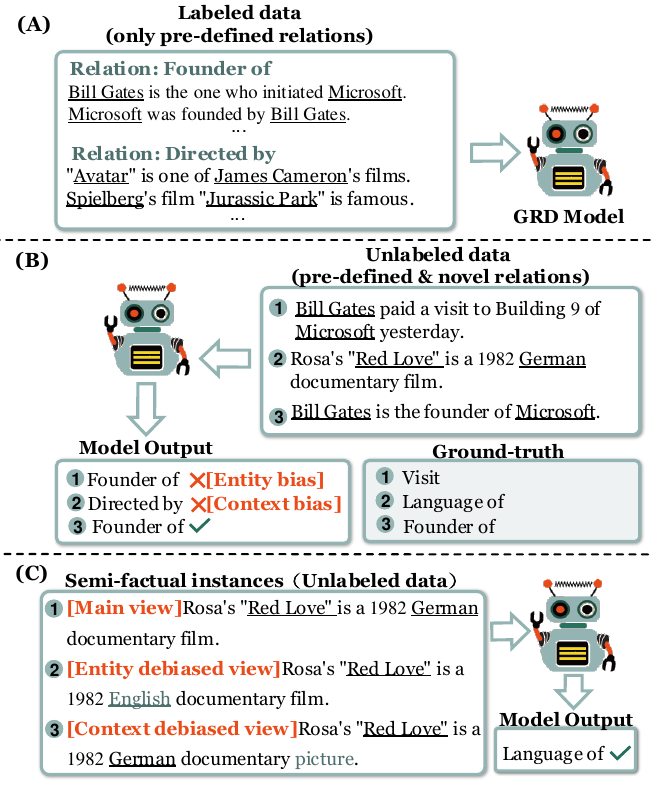} 
	\caption{Illustrative examples of entity bias and context bias as well as semi-factual instances. (A) Training on labeled pre-defined relations leads to inherent model bias; (B) Models may misidentify relations in unlabeled data due to entity and context biases; (C) Generate semi-factual instances through different debiased views. }
	\label{intro}
\end{figure}
However, this task formulation has two limitations: one is that it clusters instances without considering the meaning of each cluster, which does not satisfy the purpose of RE. Another is that it assumes the open world only contains undiscovered novel relations, which is not realistic. A more common case is that some labeled instances of pre-defined relations are accessible and the open-world unlabeled data are mixed with both pre-defined and novel relations. To this end, we introduce a new task termed {\bf G}eneralized {\bf R}elation {\bf D}iscovery (GRD) for open-world relation extraction. Under GRD, we are provided with a labeled dataset and an unlabeled dataset. In the labeled dataset, instances belong to pre-defined relations, while in the unlabeled dataset, instances belong to both pre-defined relations and an unknown number of novel relations. Addressing the GRD task involves not only classifying unlabeled instances into one of the pre-defined relations or discovering new relations to form clusters but also providing specific relational meanings for these newly discovered clusters.

Based on the above, the critical challenges for GRD are two-fold:
\begin{enumerate*}[label=(\roman*)]
\item {\bf How to mitigate the serious model biases caused by labeled pre-defined relations in GRD, and learn effective relational representations for unlabeled data?} Previous studies~\cite{DBLP:conf/naacl/WangCZCLLYLH22,DBLP:conf/sigir/HuHZKY23} show that entity and context information in training data can bias RE models (i.e., entity bias and context bias), leading to incorrect predictions or biased learning in test data. In other words, the model may predict relation only by the consistency of entity pairs or the similar contextual content, without deeply understanding the relation between entity pairs in the contexts. Notably, these biases will become more serious in the GRD task. It is mainly due to this task requiring the model to learn from labeled data with pre-defined relations, as well as unlabeled data with both pre-defined and novel relations simultaneously. As illustrated in Fig.~\ref{intro}(A) and (B), supervised signals from labeled data bias the model towards familiar, pre-defined relational patterns when handling unlabeled data. This leads to erroneous confusion between novel and pre-defined relations. \item {\bf How can we determine the specific meanings of novel relations during the classification or clustering of unlabeled instances?} Besides assigning unlabeled instances to clusters or mapping them to pre-defined relations, GRD also requires learning the semantics of clusters, i.e., knowing what the new relations are. However, classification or clustering is about learning a metric space to measure similarity between instances, so it is not easy to determine the semantic meaning of clusters in such a space. Although some OpenRE methods use Masked Language Models (MLMs) to extract representations that may contain some semantics~\cite{DBLP:conf/emnlp/LiJH22}, this information will be diluted when applied to a metric space for further processing.

\end{enumerate*}

 In this article, we propose SFGRD, a novel framework that learns from {\bf S}emi-{\bf F}actuals to address the abovementioned problems for {\bf G}eneralized {\bf R}elation {\bf D}iscovery. Semi-factuals provide contrastive explanations to facilitate semi-factual thinking for a better understanding of facts~\cite{DBLP:conf/aaai/KennyK21,DBLP:conf/acl/LuYMZ22}, i.e., even if some contents in the original text were changed, the underlying focus remained the same. This inspired us to modify bias-prone parts of sentences, generating semi-factual examples and guiding the model to do semi-factual thinking based on these examples, thereby reducing bias and promoting relation learning. As illustrated in Fig.~\ref{frema}, SFGRD consists of two main stages: the first is semi-factual generation, implemented through the tri-view debiased relation representation module, and the second is semi-factual thinking, executed by the dual-space tri-view collaborative relation learning module. In the first stage, we regard each original sentence as the main view and design two debiased views to generate semi-factual contrastive sentences: one by replacing head and tail entities with type names, and another by substituting context words with synonyms. We then construct prompts to initialize relation representations of these views via MLMs, which provide natural conditions for obtaining relational semantics. In the second stage, to preserve semantic information in metric learning, we introduce a cluster-semantic space for mainly consolidating relational semantics and a class-index space for primarily learning relation label indices. The former refines relational semantics by fine-tuning MLMs through self-contrastive learning, the latter adopts a classifier to categorize unlabeled instances into known relations heads and activate other heads to identify new ones. To enable semi-factual thinking across three views and establish a closed self-supervised learning loop for unlabeled data, we devise alignment and selection strategies to make two spaces work collaboratively. Alignment synchronizes tri-view clustering results by anchor labels from the class-index space and assigns cluster labels to each instance. Selection evaluates three views in the cluster-semantic space to select reliable cluster labels for feature learning in the class-index space. Finally, the class-index space outputs label indices, and the cluster-semantic space generates specific meanings for each class.
 
 To sum up, three main contributions of this work are summarized as follows.
 
 \begin{enumerate}
 	\item We formally define the Generalized Relation Discovery (GRD) task, which consists of identifying open-world unlabeled instances into one of pre-defined relations or discovering novel relations to form clusters and providing specific meanings for these discovered novel relation clusters. This task definition is novel and realistic for open-world relation extraction.
 	\item We highlight two challenges of the GRD task. On the one hand, it needs to mitigate model biases caused by labeled pre-defined relations and learn effective relational representations for unlabeled data. On the other hand, it should determine the specific semantics of novel relations during classifying or clustering unlabeled instances.
 	\item We propose a novel framework SFGRD for the GRD task. SFGRD first generates semi-factuals and then engages semi-factual thinking by the collaboration of cluster-semantic space and class-index space, thereby mitigating biases and promoting semantic-aware relation learning. To our best knowledge, we are the first to exploit the efficacy of semi-factuals in relation extraction.
 	
 	\item We set up benchmarks for SFGRD performance evaluation. Extensive experimental results on two datasets demonstrate that SFGRD surpasses state-of-the-art baseline models in terms of accuracy by 2.36\% $\sim$5.78\% and cosine similarity by 32.19\%$\sim$ 84.45\% for relation label index and relation semantic quality, respectively.
 \end{enumerate}

\section{Related work}
\label{sec:related}


\subsection{Semi-factual Thinking}
Semi-factual thinking is a common form of human explanation and has been researched in psychology for decades~\cite{mccloy2002semifactual}. Similar to counterfactuals, semi-factuals offer contrastive explanations for understanding facts~\cite{roese1997counterfactual, DBLP:conf/ijcai/AryalK23}. However, different from counterfactuals that alter all causal terms to obtain different outcomes, semi-factuals change conditions and keep the outcome unchanged. Therefore, semi-factuals are easier to generate than counterfactuals because they require fewer feature changes. Actually, explanations that require fewer meaningful changes to the original are often more intuitive to comprehend~\cite{DBLP:conf/aaai/KennyK21}.  As a result, semi-factual thinking has received a growing number of studies in recent years. Specifically, KennyK et al.~\cite{DBLP:conf/aaai/KennyK21} proposed an algorithm using counterfactuals and semi-factuals for image classification. Lu et al.~\cite{DBLP:conf/acl/LuYMZ22} generated semi-factuals to decouple spurious associations for sentiment classification. Vats et al.~\cite{DBLP:conf/icassp/VatsMPW23} found semi-factual explanations for classifications of medical images. Taking inspiration from the above success, we first introduce semi-factual thinking to relation learning, which creatively designs our model based on this principle to mitigate biases and achieve better relation learning.

\subsection{Open Relation Extraction}
Relation Extraction (RE) is one of the most essential technologies for knowledge graph construction. Traditional RE methods mainly focus on identifying the relations between two entities from pre-defined relation categories~\cite{DBLP:conf/emnlp/DongPL21,DBLP:conf/emnlp/LiLDYLH21, DBLP:conf/coling/JiYLMWTL20} and are incapable of extracting novel relations emerging in the real world. Open relation extraction (OpenRE) has been explored to meet this needs. Previous work can be divided into tagging-based methods~\cite{DBLP:conf/ijcai/BankoCSBE07,DBLP:conf/naacl/YatesBBCES07} and clustering-based methods~\cite{DBLP:conf/emnlp/HuWXZY20, DBLP:conf/acl/SimonGP19,DBLP:conf/esws/ElSaharDGGL17}. Tagging-based methods usually extract multiple phrases in sentences as relations and lack generality due to the sentences with the same relation may be expressed differently~\cite{DBLP:conf/emnlp/FaderSE11}. Comparatively, clustering-based methods have drawn more attention. Zhang et al.~\cite{DBLP:conf/naacl/ZhangYXHLLLS21} integrated hierarchy information into relation representations for better novel relation clustering. Liu et al.~\cite{DBLP:conf/acl/LiuYLH020} revisited OpenRE from a causal view and formulated relation clustering by a structural causal model. Zhao et al.~\cite{DBLP:conf/emnlp/ZhaoGZZ21} adopted large amounts of pre-defined relational instances to learn a relational-oriented clustering model for novel relation clustering. However, all these methods assume that the open world contains only undiscovered novel relations and just do clustering without considering the specific meaning of each cluster. Therefore, we propose the Generalized Relation Discovery (GRD) task that aims to identify unlabeled instances into a pre-defined relation or discover novel relations by not only assigning instances to clusters but also providing specific meanings for these clusters. 

\subsection{Generalized Category Discovery}
The setting we propose for GRD is inspired by Generalized Category Discovery (GCD), which was first proposed by Vaze et al. to predict both known and novel classes from a set of unlabeled images~\cite{DBLP:conf/cvpr/VazeHVZ22}. Concurrently, Cao et al.~\cite{DBLP:conf/iclr/CaoBL22} tackled a similar setting under the name of open-world semi-supervised learning for image recognition. Aside from the difference in domains, their setting differs from ours in that they only learn the category index information for open-world unlabeled data. In other words, they only classify or cluster unlabeled data into different groups but do not care about their meanings. Since the purpose of RE is to identify relations between entities, it is also desirable to learn the specific meaning of relations in unlabeled data, in addition to learning only category index information. Thus, our proposed GRD task aims not only to determine class indices but also to learn specific meanings for each class.


\section{Problem Formulation}
We first formalize the GRD as follows. Define a relational instance as a tuple: $x = \left \langle \bm s,h,t\right\rangle $, where $\bm s = \{s_i\}_{i=1}^{n}$ represents a sentence with $n$ tokens, $h$ and $t$ are the head entity and tail entity in $\bm s$, respectively. According to whether they have annotations, real-world relational instances can be categorized into labeled and unlabeled sets, denoted as \(X^l=\{x_i^l\}_{i=1}^{N_1}\) and \(X^u=\{x_i^u\}_{i=1}^{N_2}\). For the labeled instance set $X^l$, it conveys relations in a pre-defined set $R^l$ and has a corresponding label index set $Y^l=\{y_i^l\}_{i=1}^{N_1}$. Here, every $y_i^l$ is an index and can be mapped to a specific relation in $R^l$. For the unlabeled instance set $X^u$, we assume it ground-truth relation set as $R^u$. Note that $R^u=R^l \cup R^n$, where $R^l$ and $R^n$ are the pre-defined and novel relations, respectively. The objective of GRD includes two aspects: one is to predict the label index set $Y^u=\{y_i^u\}_{i=1}^{N_2}$ for $X^u$, and another is to provide the specific relational words for newly discovered novel relations $R^n$ in $X^u$. We refer to $R^l$, $R^n$, and $R^u$ as ``Pre-defined'', ``Novel'', and ``All'' relations, respectively. They are abbreviated as ``Pre'', ``Nov'', and ``All''.

\begin{figure*}
	\centering
	\includegraphics[width=1\linewidth]{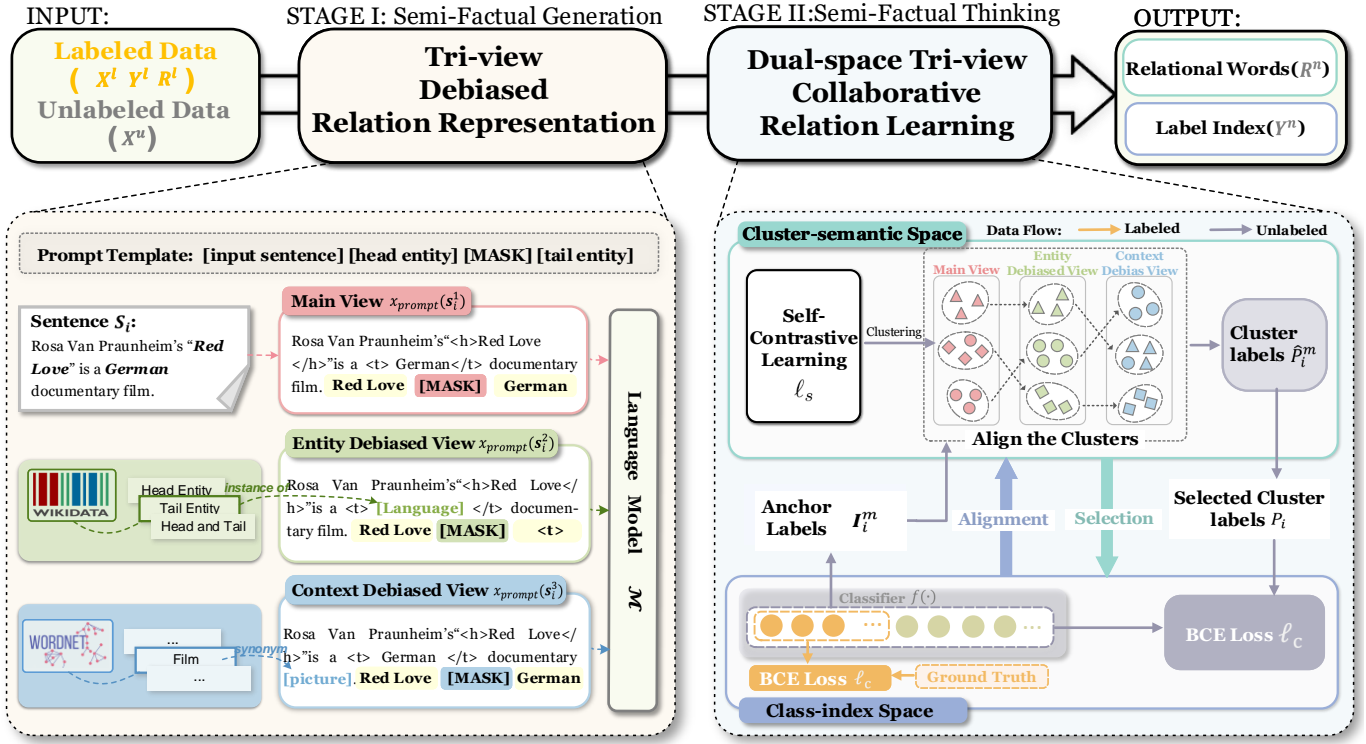} 
	\caption{Overview of SFGRD. The input consists of labeled and unlabeled data, with labeled data only containing pre-defined relations, and unlabeled data including both pre-defined and novel relations. After two-stage learning, SFGRD outputs label indices and relational words for unlabeled data.}
	\label{frema}
\end{figure*}

\section{Methodology}

 
As illustrated in Fig.~\ref{frema}, SFGRD is a two-stage framework, including semi-factual generation and semi-factual thinking. The first stage is implemented through the {\bf tri-view debiased relation representation module}. We take the original example as the main view and design two biased views to generate two semi-factual examples that express the same relation as the original one. The second stage is achieved by the {\bf dual-space tri-view collaborative relation learning module}. We design a cluster-semantic space to consolidate relational semantics and a class-index space to learn relation class indices. To facilitate collaborative work between these two spaces and integrate information from tri-views to achieve semi-factual thinking, we devise alignment and selection strategies for relation learning. The details of stage one and stage two are introduced in Section~\ref{relation Repres} and Section~\ref{stage two} respectively. The optimization and inference procedures are described in Section~\ref{opt} and summarized in Algorithm~\ref{algo}.

\subsection{ Tri-view Debiased Relation Representation }
\label{relation Repres}
	

{\bf Main View. } Similar to~\cite{DBLP:conf/acl/SoaresFLK19}, we mark the entity pair in sentence $\bm s_i$ with special tokens $\langle h \rangle$, $\langle / h \rangle$, $\langle t \rangle$, and $\langle /t \rangle$, respectively. This view keeps the whole information of the original sentence and additionally emphasizes the position of the head and tail entities in sentences.
	
{\bf Entity Debiased View.} We first use BLINK~\cite{DBLP:conf/emnlp/WuPJRZ20} to link $h_i$ and $t_i$ to Wikidata and query the Wikidata statement under \texttt {InstanceOf} status. This statement provides generalized characteristics for input entities, e.g., the results for ``German'' and ``Beijing'' are ``[Language]'' and ``[City]'', respectively. Next, we adopt these results to randomly replace $h_i$, or $t_i$, or both in sentence $\bm s_i$ to mitigate entity bias. 
	
	
	
{\bf Context Debiased View. } We substitute words in the context with synonymous terms and maintain semantic relevance and syntactic correctness to alleviate context bias. These synonyms can be generated manually by humans, automatically through a lexicon, e.g., WordNet~\cite{DBLP:journals/cacm/Miller95}, or by utilizing a pre-trained language model~\cite{DBLP:conf/naacl/DevlinCLT19}. For simple and effective, we utilize WordNet to generate synonyms and randomly replace 5\% of the context words. Note that we use Stanford CoreNLP~\cite{DBLP:conf/acl/ManningSBFBM14} for part-of-speech parsing, preventing substitution of proper nouns, pronouns, coordinating conjunctions, etc.

{\bf Relation Representation. } Formally, for one sentence $\bm s_i$, $i \in [1, N_1+N_2]$, we use $\bm s^{m}_i$, $m \in [1,3]$ to represent this sentence processed by the above three views, where $m=1$, $m=2$, and $m=3$ denote the main view, entity debiased view, and context debiased view, respectively. Note that the relation label of $\bm s^m_i$ is always the same as the original sentence $\bm s_i$. Next, we adopt the prompt-based method to learn relations through $[${MASK}$]$ tokens. Specifically, given a relational instance at the $m$-th view $x^m_i = \left \langle \bm s^m_i,h_i,t_i\right\rangle $, we first construct a template $x_{\text {prompt}}(\cdot) =$ ``$[h]$ $[$MASK$]$ $[t]$'' and map $x^m_i$ into $x_{\text {prompt}}(\bm s^m_i)$ $=$ $\bm s^m_i$ $\oplus$ $h_i$ $[$MASK$]$ $t_i$, where $\oplus$ is the string concatenation operation. In the entity debiased view ($m=2$), however, the replaced head or tail entities also should not appear in the template, so we use special tokens $\langle h \rangle$ and $\langle t \rangle$ to represent them, e.g.,  as shown in Fig.~\ref{frema}, we construct a template ``$x_{\text {prompt}}(\cdot) =$ $\text {Red Love}$ $[$MASK$]$ $\langle \text{t}\rangle$'' for this view. Then we feed $x_{\text {prompt}}(\bm s^m_i)$ to a language model ${\mathcal M}$ as:

\begin{equation}
		\bm x^m_{i,[\text {MASK}]}, \bm v^m_{i,[\text {MASK}]} = \mathcal M(x_{\text {prompt}}(\bm s^m_i))
		\label{rep1}
\end{equation}
where $\bm x^m_{i,[\text {MASK}]}$ and $\bm v^m_{i,[\text {MASK}]}$ are the hidden vector and vocabulary distribution in $[${MASK}$]$ token of $m$-th view for $i$-th instance, respectively. We take $\bm x^m_{i,[\text {MASK}]}$ and $\bm v^m_{i,[\text {MASK}]}$ as relational hidden vector $\bm x^m_i$ and relational word distribution $\bm v^m_i$, respectively.

\subsection{Dual-Space Tri-view Collaborative Relation Learning}
\label{stage two}
	
After obtaining the initialized relational hidden vector and word distribution for each input instance, we propose a dual-space tri-view collaborative relation learning module to detect relations in unlabeled data. The module details are as follows.

{\bf Cluster-Semantic Space.} In this space, we first refine the relational word distribution $\bm v^m_i$ across tri-view by using self-contrastive learning~\cite{DBLP:journals/tkde/LiuZHMWZT23} to fine-tune $ \mathcal M$. Specifically, each $\bm v^m_i$ associates with $(3N_2-1)$ examples, we treat $v^n_i, n \neq m, n\in[1,3]$ as positive examples and the rest $3(N_2-1)$ as negative examples. Then for each $\bm v^m_i$, we push it closer with positive examples and separate it apart from all negative examples by minimizing the following: 	

\begin{equation}
	{\ell_s} =  - \frac{1}{{{N_2}}}\sum\limits_{i = 1}^{{N_2}} {\log \frac{{\sum\nolimits_{n \ne m,n = 1}^3 {\exp ({{\text{sim}(\bm v_i^m,\bm v_i^n)} \mathord{\left/
							{\vphantom {{\text{sim}(\bm v_i^m,\bm v_i^n)} {{\tau_1}}}} \right.
							\kern-\nulldelimiterspace} {{\tau _1}}})} }}{{\sum\nolimits_{j = 1}^{{N_2}} {\sum\nolimits_{u = 1}^3 {\exp ({{\text{sim}(\bm v_i^m,\bm v_j^u)} \mathord{\left/
								{\vphantom {{\text{sim}(\bm v_i^m,\bm v_j^u)} {{\tau _1}}}} \right.
								\kern-\nulldelimiterspace} {{\tau _1}}})} } }}} ,
	\label{e2}
\end{equation}
where $\text{sim}(\cdot,\cdot)$ computes the similarity between the example pair and we denote it as a dot product for simple and effective. $\tau_1$ is the temperature parameter. Next, we adopt K-Means~\cite{DBLP:conf/soda/ArthurV07} to obtain the representations $\bm \mu_c^m, c\in\{1,\ldots,C\}$for $C$ cluster centroids in $m$-th view (The estimation of the cluster number is in Section~\ref{cluster_num}). Following~\cite{DBLP:conf/naacl/ZhangNWLZMNAX21, DBLP:journals/tip/XiaWGYG23}, we leverage the Student's t-distribution with one degree to define the similarity between each instance and every cluster centroid as:
\begin{equation}
	{p^m_{ic}} = \frac{{{{(1 + \left\| {\bm v_i^m - {\bm \mu _c^m}} \right\|_2^2)}^{ - 1}}}}{{\sum\nolimits_{c' = 1}^C {{{(1 + \left\| {\bm v_i^m - {\bm \mu _{c'}^m}} \right\|_2^2)}^{ - 1}}} }},
	\label{a1}
\end{equation}
where $p^m_{ic}$ represents probability of assigning the $i$-th instance on $m$-th view to the cluster center $c$.
Afterwards, we obtain the cluster label $P_i^m$ for every instance in each view by:
\begin{equation}
	P_i^m = \mathop {\arg \max }\limits_{c\in\{1,...,C\}} ({p_{ic}^m}),
	\label{a2}  
\end{equation}

{\bf Class-Index Space.} In this space, we first obtain label index distribution from hidden vectors $\bm x^m_i$ via a shared classifier $f(\cdot)$, denoted as ${\bf z}_{i}^m = f(\bm x^m_i)$. On this basis, the probability of the $i$-th instance belongs to the $j$-th class in the $m$-th view is $z_{i,j}^m \in {\bf z}_{i}^m, j\in \{1,\cdots,K\}$. Here, $K= |R^u|=|R^l|+|R^n|$, representing the total number of classification heads, which includes both pre-defined and novel relations. We set $K$ equal to the cluster number estimated in the cluster-semantic space, i.e., $K=C$. To simplify the notation, we use $C$ to represent $K$ in subsequent formulas. Considering that different views of the same example always convey the same relation, we preserve the consistency of label index assignments across tri-view by:
\begin{equation}
	{\ell_p} =  - \frac{1}{C}\sum\limits_{j = 1}^C{\log \frac{{\sum\nolimits_{n \ne m,n = 1}^3 {\exp ({{\text{sim}(z_{:,j}^m,z_{:,j}^n)} \mathord{\left/
							{\vphantom {{\text{sim}(z_{:,j}^m,z_{:,j}^n)} {{\tau _2}}}} \right.
							\kern-\nulldelimiterspace} {{\tau _2}}})} }}{{\sum\nolimits_{c= 1}^C {\sum\nolimits_{u = 1}^3 {\exp ({{\text{sim}(z_{:,j}^m,z_{:,c}^u)} \mathord{\left/
								{\vphantom {{\text{sim}(z_{:,j}^m,z_{:,c}^u)} {{\tau _2}}}} \right.
								\kern-\nulldelimiterspace} {{\tau _2}}})} } }}} - H(Z),
	\label{consis}
\end{equation}
where $z_{:,j}^m$ represents label assignment of the $m$-th view on the $j$-th class and $\tau _2$ is the temperature parameter. $H(Z)$ is a regularization term~\cite{DBLP:conf/acl/MouHWZXJWX22} that prevents the simplistic solution where most instances are assigned the same label. Moreover, $H(Z)$ is computed as the entropy of the label assignment probabilities, averaged over all instances, i.e., $H(Z)=\sum\nolimits_{m = 1}^3 {\sum\nolimits_{j = 1}^C {Z_j^m\log Z_j^m} }$, where $Z_j^m = \frac{1}{{{N_2}}}\sum\nolimits_{i = 1}^{{N_2}} {z_{i,j}^m}$.

{\bf Collaborative for Relation Learning.}	
To enable semi-factual thinking across three views and establish a closed self-supervised learning loop for unlabeled data relation learning, we devise alignment and selection strategies to make two spaces work collaboratively. The details are as follows.

\subsubsection{Alignment strategy}
The consistency labels of three views in the class-index space can be used to guide the alignment of the cluster results for different views in the cluster-semantic space. Formally, the labels for $i$-th instance of $m$-th view in class-index space is: 
\begin{equation}
	I_i^m = \mathop{\arg \max }\limits_{j\in\{1,\ldots,C\}} ({z_{i,j}^m}) ,
	\label{a3}
\end{equation}
according to Equation~\ref{consis} in advance, the predicted labels for the same instance in different views are the same, i.e., $I_j^m=I_j^n, m\neq n$.  Therefore, we treat $\{I^m_i\}_{i=1}^{N_2}$ as anchor labels to modify $\{s_i^m\}_{i=1}^{N_2}$ by:
\begin{equation}
	\begin{array}{l}
		\mathop {\min }\limits_{{\bm U^m}} {\bm Q^m}{\bm U^m}\\
		= \mathop {\min }\limits_{{\bm U^m}} \left[ {\begin{array}{*{20}{c}}
				{q_{11}^m}& \cdots &{q_{1C}^m}\\
				\vdots & \ddots & \vdots \\
				{q_{C1}^m}& \cdots &{q_{CC}^m}
		\end{array}} \right]\left[ {\begin{array}{*{20}{c}}
				{u_{11}^m}& \cdots &{u_{1C}^m}\\
				\vdots & \ddots & \vdots \\
				{u_{C1}^m}& \cdots &{u_{CC}^m}
		\end{array}} \right]\\
		\\
		\begin{array}{*{20}{l}}
			{s.t.}&{q_{cc'}^m = \mathop {\max }\limits_{c,c'} (\hat q_{cc'}^m) - \hat q_{cc'}^m,}\\
			{}&{\sum\limits_{c = 1} {u_{cc'}^m = 1,\sum\limits_{c' = 1} {u_{cc'}^m = 1,} } }\\
			{}&{c,c' = 1,2, \ldots ,C},
		\end{array}
	\end{array}
	\label{Ha}
\end{equation}
where $\bm Q^m$ and $\bm U^m$ both with a dimension of $C \times C$ represent the cost matrix and boolean matrix, respectively. $\hat q_{ij}^m = \sum\limits_{i = 1}^{N_2} {\mathbb{I}(I_i^m = c)} \mathbb{I}(P_i^m = c)]$ and $\mathbb{I}(\cdot)$ denotes the indicator function. We optimize Equation~(\ref{Ha}) the Hungarian algorithm~\cite{jonker1986improving}. Then, for the $i$-th instance of the $m$-th view, if its original cluster label $ P_i^m=c'$ satisfies equation~(\ref{cluster}), its aligned cluster label can be denoted as $\hat P_i^m=c$.
\begin{equation}
	c = c \cdot (\mathbb{I}(u^m_{cc'}=1)\mathbb{I}( P_i^m=c')), c,c'\in\{1,\ldots,C\}.
	\label{cluster}
\end{equation}	

\subsubsection{Selection strategy}

The aligned cluster labels of different views can be used to guide feature learning in the class-index space. However, despite Equation~(\ref{Ha}) aligning the cluster labels of most instances across three views, some instances are still inconsistent among the three views. To reduce model confusion during training, we expect the model to be able to perform semi-factual thinking, that is, to select reliable cluster labels by integrating information from tri-view and dual-space. Thus, we devise the following selection strategy:
\begin{equation}
	\mathcal{P}_i = \left\{ {\begin{array}{*{20}{l}}
			{\hat P_i^m,}&{\text{if}\ \forall m,n \in \{ 1,2,3\} ,\hat P_i^m = \hat P_i^n;}\\
			{}&{}\\
			{\hat P_i^g,}&{\text{if}\ \exists m,n,g \in \{ 1,2,3\} ,\hat P_i^m \ne \hat P_i^n,}\\
			{}&{\text{and}\ \mathop {\max }\limits_m \hat p_{ic}^m = \hat p_{ic}^g \ge \theta ;}\\
			{abandon,}&{\text{if}\ \exists m,n,g \in \{ 1,2,3\} ,\hat P_i^m \ne \hat P_i^n,}\\
			{}&{\text{and}\ \mathop {\max }\limits_m \hat p_{ic}^m = \hat p_{ic}^g < \theta ;}
	\end{array}} \right.
	\label{sele}
\end{equation}

where $\hat p_{ic}^m$ represents the probability of assigning the cluster label to the $i$-th instance in class $c$, and $\theta$ denotes a threshold. Concretely, this formula operates to select the reliable cluster label by considering the following three cases: 
\begin{itemize}[nosep,leftmargin=*]
	\item First, when the aligned cluster labels across all views are equal, we use $\hat P_i^m$ as the cluster label directly;
	\item Second, if the aligned cluster labels are different across views, we conduct a comparison of the cluster assignment probabilities among three views, and obtain the maximum value $\hat p_{ic}^g$. If this value exceeds the threshold $\theta$, the label in view $g$ is selected as the final cluster label, i.e., $\hat P_i^g$.
	\item Third, if $\hat p_{ic}^g$ fails to exceed $\theta$, we abandon this instance. 
\end{itemize}
After selecting the reliable cluster labels, we then fine-tune our model by both labeled and unlabeled instances as:
\begin{equation}
	{\ell_c} =  - \frac{1}{3N}\sum\limits_{m = 1}^3\sum\limits_{i = 1}^N CE(Y_i,z_{i,:}^m)
	\label{e3}
\end{equation}

where $CE(\cdot,\cdot)$ is the cross entropy loss function and $z_{i,:}^m$ represents a probability distribution over $C$ classes for the $i$-th instances in class-index space. Note that for $i$-th instance $x_i \in \mathcal{D}^{u}$, $Y_i = \mathcal{P}_i$ and $N= N_2$; For $i$-th instance $x_i \in \mathcal{D}^{l}$, $Y_i$ is the ground-truth label and $N= N_1$. 
		
\begin{algorithm}[t]	
	\SetAlgoNoLine
	\SetKwInOut{Input}{\textbf{Input}}\SetKwInOut{Output}{\textbf{Output}} 	
	\Input{Labeled dataset $\mathcal{D}^{l}$ and unlabeled dataset $\mathcal{D}^{u}$; temperature parameter $\tau _1$ and $\tau _2$; threshold $\theta$; number of classification heads and cluster centers $C$; language model $\mathcal M$; classifier $f(\cdot)$.  }
		
	\Output{ Optimized model parameters \;\\
		}
	\BlankLine
	\tcp{Tri-View debiased relation representation}
	\For{$i=1 \rightarrow N_1+N_2$}{
			Generate semi-factual examples for the input instance $\bm s_i$ and obtain $\bm s_i^m,m\in\{1,2,3\}$\;}
	\For{$i=1 \rightarrow N_1+N_2$}{Obtain the hidden vector $\bm x_i$ and relational word distribution $\bm v_i^m$ by Eq.~(\ref{rep1})\;}
	\tcp{Dual-Space Tri-View Collaborative Relation Discovery}
		Initialize $\mathcal M$ and $f(\cdot)$ by Eqs.~(\ref{e2}), (\ref{consis}) and (\ref{e3}) with instances in $\mathcal{D}^{l}$\;
	\While{{\texttt {SFGRD}} does not converge}{
			For instances in $\mathcal{D}^{u}$, refine relational semantics in cluster-semantic space by Eqs.~(\ref{e2})\;
			Obtain cluster labels by Eqs.~(\ref{a1}) and (\ref{a2})\;
			Learn anchor labels in class-index space by Eq.~(\ref{a3}) and Eq.~(\ref{consis})\;
			Align cluster labels across different views by solving Eq.~(\ref{Ha}) and obtain the aligned cluster labels by Eq.~(\ref{cluster}) holds \;
			Select reliable cluster labels by Eq.~(\ref{sele})\;
			Fine-tune $\mathcal M$ and $f(\cdot)$ by Eq.~(\ref{e3})\;	
			Update model parameters in the next episode.
		}	
	\Return Model parameters of {\texttt {SFGRD}}.	
	\caption{The Learning Process of {\texttt {SFGRD}}.}\label{algorithm1}
	\label{algo}
\end{algorithm}

\subsection{Optimization and Inference.} 
\label{opt}
We summarize the overall learning process of {\texttt {SFGRD}} in Algorithm~\ref{algo}. Specifically, we take the original relational example as the main view and generate two semi-factual examples that express the same relation as the original one from two debiased views. Next, we obtain the initialized hidden vector and word distribution for each instance. Then, we refine relational semantics in cluster-semantic space by Eqs.~(\ref{e2}) and adopt self-contrastive learning in class-index space to learn anchor labels by Eq.~(\ref{consis}) and Eq.~(\ref{a3}). Afterwards, the cluster labels obtained by Eqs.~(\ref{a1}) and (\ref{a2}) are then aligned through the maximum matching formula in Eq.~(\ref{Ha}). Finally, we perform Eq.~(\ref{sele}) to select the reliable cluster labels and fine-tune our model by Eq.~(\ref{e3}). At the inference stage, the class-index space determines the label index for the $i$-th instance by ${y_i} = \mathop {\arg \max }\limits_{j \in { 1,...C} } (\frac{1}{3}\sum\limits_{m = 1}^3 {z_{i,j}^m} )$, while concurrently, the cluster-semantic space outputs relational words relevant to this instance.

\section{Experiments}
\label{sec:exp}
\label{exps}
In this section, we describe the experimental settings, which include datasets in Section~\ref{dataset}, comparison models in Section~\ref {comM}, as well as the implementation and evaluation metrics in Section~\ref{Imple}. Then, we evaluate the overall performance of SFGRD in Section~\ref{MianRes} followed by ablation studies in Section~\ref{Ablation} and do the relation number estimation in Section~\ref{cluster_num}. Finally, we discuss the correlation between label index accuracy and semantic quality, along with some results for predicted relational words in Section~\ref{case}.

\begin{table}[t]
	\footnotesize
	\centering
	\caption{Dataset statistics.}
	\label{tab:statictics}
	\setlength{\tabcolsep}{1.8mm}{
		\begin{tabular}{ccc}
			\toprule
			\bf Dataset                         & \bf FewRel           & \bf TACRED         \\ \midrule
			\# All Relations                & 80               & 41             \\ 
			\# Pre-defined Relations        & 64/40/16         & 32/21/9        \\ 
			\# Novel Relations              & 16/40/64         & 9/20/32        \\ 
			\# Labeled examples             & 4800/12000/19200 & 1052/1944/4132 \\ 
			\# Unlabeled examples           & 24000            & 6341           \\ 
			\#Test examples(Test Relations) & 8000(80)         & 2218(41)       \\ \bottomrule
	\end{tabular}}
\end{table}

\subsection{Datasets}
\label{dataset}

We evaluate our model on two public relation extraction datasets: FewRel~\cite{DBLP:conf/emnlp/HanZYWYLS18} and TACRED~\cite{DBLP:conf/emnlp/ZhangZCAM17}. The statistical results of the datasets are shown in Table~\ref{tab:statictics}. 
\begin{enumerate}
\item {\bf FewRel} is derived from Wikipedia and contains 70,000 instances in 100 relations, that is, each relation has 700 instances. However, only 80 relations have been released.  For the GRD task, we set the number of all relations $R^u$ as 80, and adjust the number of novel relation $R^n$ and pre-defined relation $R^l$ with different novel relation ratios. For the test set, we randomly select 100 examples from 80 relations, i.e., 8000 examples. For the unlabeled data $\mathcal D^u$, we select 50\% of the samples from 80 relations of the remaining examples, i.e., 80 relations and each with 300 examples. For the labeled data $\mathcal{D}^l$, the number of pre-defined relations varies with the current novel relation ratio and we take another 300 examples for each relation. 

\item {\bf TACRED} is a human-annotated dataset. It is built over newswire and web text from the corpus used in the yearly TAC Knowledge Base Population(TAC KBP) challenges. It contains 42 relations, we remove the examples labeled as {\em No relation} and use the remaining 41 relations for our experiments. We set the number of all relations $R^u$ as 41, and adjust the number of novel relation $R^n$ and pre-defined relation $R^l$ with different novel relation ratios. For the test set, we randomly select 15\% of the examples from all relations. For the unlabeled data $\mathcal D^u$, we select 50\% of the remaining examples for 41 relations. For the labeled data $\mathcal{D}^l$, the number of pre-defined relations varies with the current novel relation ratio and we take another 50\% of the examples for each relation. 
\end{enumerate}


\subsection{Comparing Methods}
\label{comM}
We compare SFGRD with nine outstanding baselines to demonstrate its superiority. These baselines can be categorized into three groups.
\begin{enumerate*}[label=(\roman*)]
	\item Five open relation extraction methods: SelfORE~\cite{DBLP:conf/emnlp/HuWXZY20}, RoCORE~\cite{DBLP:conf/emnlp/ZhaoGZZ21}, CaPL~\cite{DBLP:conf/coling/DuanWLX22}, MatchPrompt~\cite{DBLP:conf/emnlp/WangZLLZW22}, and TABs~\cite{DBLP:conf/emnlp/LiJH22}. To match the original training pattern of these models, we mix pre-defined relations with novel relations, and make models treat all these relations as ``novel''.
	
	\item Three typical semi-supervised clustering methods in other tasks: DeepAligned~\cite{DBLP:conf/aaai/Zhang0LL21}, GCD~\cite{DBLP:conf/cvpr/VazeHVZ22},  DCCL~\cite{DBLP:conf/cvpr/PuZS23}. To adapt these methods to our task, we obtain their source codes on Github and replace the encoder part with BERT-base for relation representation.
	
	\item The large language model: ChatGPT~\cite{chatgpt}. We use in-context learning by presenting examples from labeled data along with a list of pre-defined relation classes. Then, we instruct the model to output the pre-defined relation name or generate a novel class name when an example does not fit within the set of pre-defined relations. Afterwards, we adopt BERT-base to encode various relational phrases generated previously and use K-Means for clustering.
\end{enumerate*}
The details of baseline models are as follows.

\begin{enumerate}
	\item {\bf SelfORE}~\cite{DBLP:conf/emnlp/HuWXZY20}: generates self-supervised signals for adaptive clustering based on contextualized relational features.
	\item {\bf RoCORE}~\cite{DBLP:conf/emnlp/ZhaoGZZ21}: learns the relation-oriented representation and cluster by using a language amount of labeled data with pre-defined relations.
	\item {\bf CaPL}~\cite{DBLP:conf/coling/DuanWLX22}: a cluster-aware pseudo-labeling method to transfer more knowledge for discovering novel relations. 
	\item {\bf MatchPrompt}~\cite{DBLP:conf/emnlp/WangZLLZW22}: a prompt-based open relation clustering framework that can learn representations with efficient knowledge transfer.
	\item {\bf TABs}~\cite{DBLP:conf/emnlp/LiJH22}: a clustering method for open relation and event type discovery, which learn representations from token view and mask view.
	\item {\bf DeepAligned}~\cite{DBLP:conf/aaai/Zhang0LL21}: an unsupervised method of discovering new intents and considering the case that unlabeled data is mixed by known and new intents.
	\item {\bf GCD}~\cite{DBLP:conf/cvpr/VazeHVZ22}: the first work for generalized category discovery in computer vision.
	\item {\bf DCCL}~\cite{DBLP:conf/cvpr/PuZS23}: the currently state-of-the-art generalized category discovery method in computer vision.
	\item {\bf ChatGPT}: the excellent large language model that performed well on many natural language processing tasks, we also employ it to this task. 
\end{enumerate}
\begin{table*}[]
	\caption{Model performance with different novel relation ratios on FewRel. With 80 relations in unlabeled data, 20\%, 50\%, and 80\% correspond to 16, 40, and 64 novel relations, respectively. The best results are marked in bold, and the second best results are marked in underline.}
	\centering
	\label{FewRel}
	\renewcommand\arraystretch{1}{
		\setlength{\tabcolsep}{3.6mm}{
			\begin{tabular}{c|c|ccc|ccc|ccc}
				\toprule
				\multirow{2}{*}{Nov/All} & \multirow{2}{*}{Model} & \multicolumn{3}{c|}{ACC} & \multicolumn{3}{c|}{NMI} & \multicolumn{3}{c}{ARI} \\ \cmidrule{3-11} 
				&                        & Pre    & Nov    & All   & Pre    & Nov    & All   & Pre    & Nov    & All   \\ \hline
				\multirow{11}{*}{20\%}   & SelfORE (EMNLP2020)                & 70.32  & 53.21  & 60.82 & 74.23  & 67.02  & 73.44 & 60.87  & 50.36  & 54.25  \\
				& RoCORE (EMNLP2021)                 & \bf86.34  & 67.02  & \underline{73.67} & \bf86.73  & 76.18  & \underline{83.72} & \bf76.58  & 62.95  & \underline{66.55} \\
				& CaPL (COLING2022)                   & 84.40  & 68.06  & 72.91 & 86.11  & 77.56  & 81.75 & 75.01  & 63.21  & 66.23 \\
				& TABs (EMNLP2022)            & 84.65  & 69.45  & 71.53 & 83.43  & 78.01  & 80.60 & 74.90  & 64.35  & 65.72 \\
				& MatchPrompt (EMNLP2022)                  & 83.50  & \underline{70.92}  & 71.50 & 82.16  & \underline{79.45}  & 78.25 & 73.59  & \underline{65.76}  & 65.80 \\
				& DeepAligned (AAAI2019)            & 79.33       & 36.62       &53.46       & 81.11       &  51.29      &  72.77     &  65.09      &  25.96      &  49.13     \\
				& GCD (CVPR2022)                    & 57.20  & 40.31  & 42.17 & 70.81  & 65.23  & 63.40 & 51.27  & 43.05  & 45.60 \\
				& DCCL (CVPR2023)                  &61.20        &42.65        &45.34       & 71.54       & 67.30       & 68.10      & 53.33       &46.26        &47.20       \\
				& ChatGPT                  & 36.72  & 16.74   & 21.48 & 55.87  & 41.60  & 45.90 & 21.01  & 10.25  & 17.33 \\\cmidrule{2-11}
				& \bf SFGRD                  & \underline{85.39}  & \bf73.75  & \bf76.92 & 86.20  & \bf80.41  & \bf84.32 & \underline{75.45}  & \bf68.41  & \bf67.27 \\ 
				& \bf$\Delta$                   & $\downarrow$0.95  & $\uparrow$2.83  & $\uparrow$3.25 & $\downarrow$0.53  & $\uparrow$0.96  & $\uparrow$0.60 & $\downarrow$1.13  & $\uparrow$2.65  & $\uparrow$0.72 \\ \midrule\midrule 
				\multirow{11}{*}{50\%}   & SelfORE (EMNLP2020)                  & 69.63  & 45.79  & 55.26 & 72.50  & 56.36  & 70.91 & 57.85  & 43.60  & 51.64\\
				& RoCORE (EMNLP2021)                  & {84.10}  & 61.22  & 66.70 & \underline{85.72}  & 73.15  & \underline{77.79} & \bf75.49  & 55.39  & 63.95 \\
				& CaPL (COLING2022)                   & \underline{84.60}  & 63.26  & 67.37 & {85.65}  & 74.60  & 78.13 & {74.38}  & 59.95  & 64.86 \\
				& TABs (EMNLP2022)            & 83.57  & 63.66  & 66.32 & 82.56  & 75.20  & 77.62 & 73.72  & 60.82  & \underline{65.37} \\
				& MatchPrompt (EMNLP2022)                    & 81.84  & \underline{65.50}  & \underline{68.11} & 80.53  & \underline{76.02}  & 77.63 & 72.25  & \underline{61.24}  & 63.80 \\
				& DeepAligned(AAAI2019)            &78.14        &30.42        & 53.10      &82.19        & 49.15       & 71.60      &71.38        &25.52        &47.86       \\
				& GCD (CVPR2022)                    & 54.52  & 35.10  & 37.06 & 72.19  & 50.63  & 61.76 & 48.70  & 38.40  & 39.11 \\
				& DCCL(CVPR2023)                   &  58.33      & 40.25       &  43.06     & 72.60       &60.25        &65.62       & 51.79       & 42.03       &44.80       \\
				& ChatGPT                 & 35.31  & 13.95   & 18.01 & 54.62  & 37.50  & 42.13 & 22.36  & 9.70   & 15.20\\\cmidrule{2-11}
				& \bf SFGRD                  & \bf84.85  & \bf70.41  & \bf73.89 & \bf86.54  & \bf78.84  & \bf82.50 & \underline{75.02}  & \bf62.70   & \bf65.59 \\
				& \bf$\Delta$                & $\uparrow$0.25  & $\uparrow$4.91  & $\uparrow$5.78 & $\uparrow$0.82  & $\uparrow$2.82  & $\uparrow$4.71 & $\downarrow$0.47  & $\uparrow$1.46   & $\uparrow$0.22\\ \midrule\midrule
				\multirow{11}{*}{80\%}   & SelfORE (EMNLP2020)               & 68.16  & 38.90  & 46.40 & 71.74  & 52.10  & 68.73 & 56.05  & 40.85  & 48.97\\
				& RoCORE (EMNLP2021)                & \underline{83.33}  & 56.21  & 61.70 & \underline{84.65}  & 69.62  & 72.80& \underline{73.60}  & 49.12  & \underline{54.90} \\
				& CaPL (COLING2022)                    & 82.47  & 57.55  & 63.33 & 84.63  & 71.26  & 74.67 & 72.97  & 53.70  & 54.32 \\
				& TABs (EMNLP2022)            & 82.13  & \underline{60.72}  & \underline{63.80} & 81.17  & \underline{71.80}  & \underline{75.15} & 72.85  & \underline{55.35}  & 56.28 \\
				& MatchPrompt (EMNLP2022)                  & 82.91  & 58.16  & 62.41 & 81.52  & 68.70  & 72.91 & 70.23  & 53.20  & 54.34 \\
				& DeepAligned (AAAI2019)            & 75.31     & 32.56       &44.72       & 79.08       &  50.26      &  69.03     &  62.00      &  24.48      &  47.07     \\
				& GCD (CVPR2022)                      & 55.92  & 28.20  & 32.65 & 72.80  & 44.30  & 57.96 & 49.01  & 33.71  & 35.20 \\
				& DCCL (CVPR2023)                   & 57.23       & 33.30       & 36.05      & 73.46      & 51.43      &  59.65     & 50.58       & 37.50       & 39.26      \\
				& ChatGPT                 & 33.72  & 9.73   & 14.71 & 52.31  & 34.88  & 40.01 & 19.60  & 8.33   & 11.50 \\\cmidrule{2-11}
				& \bf SFGRD                  & \bf83.62   & \bf65.97  & \bf69.63 & \bf85.48  & \bf77.56  & \bf79.38 & \bf{73.83}  & \bf62.04  & \bf65.35 \\
				& \bf$\Delta$                 &  $\uparrow$0.29 & $\uparrow$5.25  & $\uparrow$5.83 & $\uparrow$0.83  & $\uparrow$5.76  & $\uparrow$4.23 & $\uparrow$0.23  & $\uparrow$6.69  & $\uparrow$10.45\\ \bottomrule
	\end{tabular}}}
\end{table*}
\begin{table*}[]
	\caption{Model performance with different novel relation ratios on TACRED. With 41 relations in unlabeled data, 20\%, 50\%, and 80\% correspond to 9, 20, and 32 novel relations, respectively. The best results are marked in bold, and the second best results are marked in underline.}
	\centering
	\centering
	\label{TACRED}
	\renewcommand\arraystretch{1}{
		\setlength{\tabcolsep}{3.6mm}{
			\begin{tabular}{c|c|ccc|ccc|ccc}
				\bottomrule
				\multirow{2}{*}{Nov/All} & \multirow{2}{*}{Model} & \multicolumn{3}{c|}{ACC} & \multicolumn{3}{c|}{NMI} & \multicolumn{3}{c}{ARI} \\ \cmidrule{3-11} 
				&                        & Pre    & Nov    & All   & Pre    & Nov    & All   & Pre    & Nov    & All   \\ \hline
				\multirow{11}{*}{20\%}   & SelfORE (EMNLP2020)        & 80.91  & 59.10  & 67.79  & 83.79  & 72.23  & 77.53  & 70.44  & 55.75  & 63.10 \\
				& RoCORE (EMNLP2021)         & \bf 87.31  & 77.61  & 80.74  & \bf88.65  & 80.17  & 82.02  & \underline{76.77}  & 68.33  & 71.44 \\
				& CaPL (COLING2022)          & 86.50  & 79.35  & \underline{82.59}  & 87.70  & 82.50  & \underline{84.19}  & 75.40  & 70.87  & \underline{72.16} \\
				& TABs (EMNLP2022)           & 85.90  & \underline{79.85}  & 82.05  & 86.90  & \underline{82.66}  & 83.88  & 75.03  & \underline{71.26 } & 72.04 \\
				& MatchPrompt (EMNLP2022)    & 85.27  & 78.60  & 80.16  & 86.79  & 81.18  & 81.75  & 74.24  & 69.37  & 70.80 \\
				& DeepAligned (AAAI2019)     & 79.41  & 36.03  & 55.28  & 80.71  & 49.87  & 68.24  & 64.27  & 27.07  & 38.06 \\
				& GCD (CVPR2022)             & 64.60  & 51.17  & 53.87  & 66.45  & 57.12  & 59.60  & 55.20  & 49.52  & 51.22 \\
				& DCCL (CVPR2023)            & 69.46  & 53.43  & 55.50  & 71.45  & 58.50  & 60.97  & 56.62  & 50.45  & 52.24 \\
				& ChatGPT                    & 52.62  & 19.40  & 31.72  & 63.31  & 47.93  & 54.27  & 48.05  & 15.84  & 24.06  \\ \cmidrule{2-11}
				& \bf SFGRD   & \underline{86.62}  & \bf 82.34  & \bf 84.95  & \underline{88.20}  & \bf83.20  & \bf85.86  & \bf 77.53  & \bf72.20  & \bf73.52 \\
				& \bf$\Delta$ &  $\downarrow$0.69      &  $\uparrow$2.99      & $\uparrow$2.36       &   $\downarrow$0.40     &  $\uparrow$0.54      &  $\uparrow$1.67      &  $\uparrow$0.76      &$\uparrow$0.94        & $\uparrow$1.36      \\ \midrule\midrule 
				\multirow{11}{*}{50\%}   & SelfORE (EMNLP2020)        & 79.40  & 52.80  & 64.14  & 82.61  & 68.37  & 75.30  & 69.52  & 53.59  & 60.34 \\
				& RoCORE (EMNLP2021)         & 85.13  & 74.52  & 76.83  & 85.81  & 76.45  & 79.98  & 74.42  & 62.10  & 65.02 \\
				& CaPL (COLING2022)          & 85.43  & 75.50  & 77.12  & \underline{85.91}  & 76.89  & 80.02  & 74.55  & 64.08  & 65.96 \\
				& TABs (EMNLP2022)           & \underline{85.52}  & \underline{76.08}  & \underline{78.36}  & 85.63  & \underline{77.50}  & \underline{81.26}  & \underline{75.25}  & \underline{65.53}  & \underline{67.10}\\
				& MatchPrompt (EMNLP2022)    & 83.60  & 74.33  & 75.89  & 83.87  & 76.09  & 78.82  & 72.44  & 61.30  & 63.30 \\
				& DeepAligned(AAAI2019)      & 79.05  & 32.61  & 41.56  & 78.87  & 47.03  & 57.10  & 64.92  & 21.80  & 26.29 \\
				& GCD (CVPR2022)             & 62.56  & 43.90  & 44.83  & 63.60  & 50.25  & 52.09  & 51.33  & 42.85  & 44.91 \\
				& DCCL(CVPR2023)             & 68.43  & 48.86  & 51.32  & 70.32  & 55.87  & 56.44  & 55.15  & 47.17  & 49.50 \\
				& ChatGPT                    & 50.13  & 17.55  & 26.48  & 60.82  & 44.36  & 43.28  & 44.02  & 12.65  & 21.92  \\ \cmidrule{2-11}
				& \bf SFGRD   & \bf86.13  & \bf80.03  & \bf82.80  & \bf87.84  & \bf82.63  & \bf83.15  & \bf76.65  & \bf68.40  & \bf70.91 \\
				& \bf$\Delta$ & $\uparrow$0.61       &$\uparrow$3.95        & $\uparrow$4.44       & $\uparrow$1.93       &  $\uparrow$5.13      & $\uparrow$1.89       & $\uparrow$1.40       & $\uparrow$2.87       & $\uparrow$3.81      \\\midrule\midrule 
				\multirow{11}{*}{80\%}   & SelfORE (EMNLP2020)        & 77.05  & 48.60  & 53.80  & 81.91  & 65.40  & 68.82  & 68.53  & 47.10  & 51.09 \\
				& RoCORE (EMNLP2021)         & 84.33  & 69.92  & 71.70  & 85.04  & 72.70  & 73.22  & 74.09  & 58.44  & 59.93 \\
				& CaPL (COLING2022)          & \underline{85.15}  & 72.52  & \underline{74.29}  & \underline{85.60}  & \underline{73.08}  & \underline{77.61}  & 74.64  & 62.33  & \underline{63.80} \\
				& TABs (EMNLP2022)           & 85.07  & 71.34  & 73.13  & 85.10  & 72.83  & 76.62  & \underline{74.70 } & 60.19  & 63.16 \\
				& MatchPrompt (EMNLP2022)    & 83.09  & \underline{72.80}  & 73.04  & 82.15  & 72.92  & 74.48  & 71.82  & \underline{62.50}  & 61.33 \\
				& DeepAligned (AAAI2019)     & 64.19  & 26.81  & 29.75  & 66.01  & 42.68  & 45.08  & 55.45  & 15.08  & 16.22 \\
				& GCD (CVPR2022)             & 60.67  & 36.45  & 40.52  & 62.33  & 49.70  & 51.35  & 50.69  & 39.06  & 41.43 \\
				& DCCL (CVPR2023)            & 65.75  & 40.10  & 44.35  & 66.98  & 52.27  & 54.62  & 53.80  & 42.06  & 43.29 \\
				& ChatGPT                    & 47.61  & 13.48   & 22.53  & 58.07  & 39.25  & 40.14  & 38.75  & 10.62  & 18.94 \\ \cmidrule{2-11}
				& \bf SFGRD   & \bf 86.06  & \bf77.50  & \bf 79.13  & \bf87.90  & \bf 78.63  & \bf 82.41  & \bf 76.36  & \bf 65.57  & \bf 67.73 \\
				& \bf$\Delta$ & $\uparrow$0.91       & $\uparrow$4.70       & $\uparrow$4.84       & $\uparrow$2.3       &   $\uparrow$5.55     & $\uparrow$4.8       & $\uparrow$1.66       &  $\uparrow$3.07      &  $\uparrow$3.93     \\ \bottomrule
	\end{tabular}}}
\end{table*}
\subsection{Implementation and Evaluation Metrics}
\label{Imple}
As the previous open relation discovery methods use the pre-trained language model BERT-base~\cite{DBLP:conf/naacl/DevlinCLT19} as encoder, our SFGRD also utilizes this model for fair comparison. The optimizer is  Adam~\cite{DBLP:journals/corr/KingmaB14}, in which the learning rate is $1e-4$. We set the maximum training epoch number as $100$ and test the model performance on the test set every epoch. The training will stop if there is performance no growth for $10$ consecutive epochs. All experiments are conducted using two NVIDIA 3090 with 24 GB of memory. We vary the ratio of novel relations among all relations as 20\%, 50\%, and 80\%. Table~\ref{tab:statictics} shows the corresponding relation types and example numbers in labeled data and unlabeled data. In most of our experiments, we set $\tau _1$, $\tau _2$and $\theta$ as $0.05$, $0.1$, and $0.7$, respectively.

We measure our model performance from two aspects: one is the accuracy in identifying relation labels, and another is the ability to generate semantics for novel relations. First, we adopt three widely used evaluation metrics to evaluate the performance of the model in identifying classes: Accuracy (ACC), Normalized Mutual Information (NMI)~\cite{xu2003document} and Adjusted Rand Index (ARI)~\cite{rand1971objective}. ACC is computed by the Hungarian algorithm~\cite{jonker1986improving}, which evaluates the best mapping between cluster assignments and true labels. NMI calculates the normalized measure of similarity between predicted labels and true labels. ARI evaluates the percentage of correct predicted pairs. Second, we quantify the model's ability to generate semantics by measuring the similarity of the relational word distribution between the model-generated and the ground-truth relations. Specifically, we obtain the ground-truth relational description$[$X$]$ and construct a template as: $[$X$]$ means $[$MASK$]$, then we feed this template into the pre-trained language model. In this way, ground-truth distributions of relations will be derived from word distributions at $[$MASK$]$ positions. Then, we adopt Cosine Similarity (COS)~\cite{dehak2010cosine} and Kullback–Leibler Divergence (KL)~\cite{ji2020kullback}. to evaluate the similarity of semantics. Higher COS or Lower KL values indicate greater similarity between two distributions. All results are averaged over three independent runs.

\label{sec:results-and-analysis}

\subsection{Main Results}
\label{MianRes}
To minimize the impact of different estimated cluster numbers on model performance, we assume that the number of relations in unlabeled data is known for all models in most of our experiments. In addition, we analyze the case when the number of clusters is unknown in Section~\ref{cluster_num}.

{\bf Label Index Prediction Results on All Relations.} Tables~\ref{FewRel} and \ref{TACRED} are the label index prediction results under different novel relation ratios on FewRel and TACRED datasets, respectively. We can make the following four observations:
\begin{enumerate}
	\item Overall (across all examples in ${\cal X}^u$, i.e., ``All''	), our SFGRD outperforms all comparison methods on all datasets, improving accuracy by 2.36\%$\sim$5.78\%, NMI by 0.6\%$\sim$4.8\%, and ARI by 0.72\%$\sim$10.45\%. Additionally, all model performances decrease with increasing novel relation ratios, but SFGRD declines the least. For example, when the novel relation ratio increases from 20\% to 80\% on the FewRel dataset, the accuracy of SFGRD decreases by 7.29\%, compared to a decrease of 7.73\%$\sim$14.42\% in other baselines. These results demonstrate not only the effectiveness and superiority of the proposed framework but also its robustness to novel relation ratios.
	
	\item The performance on pre-defined relations (i.e., ``Pre'') for all models is significantly higher than on novel relations (i.e., ``Nov''). This gap suggests a bias towards pre-defined relations in this task, leading to high performance on these but at the expense of novel relations. Nonetheless, our SFGRD can narrow this gap compared to other baseline models. This illustrates that the proposed model can effectively alleviate this bias.
	
	\item  Although semi-supervised clustering is closer to GRD in terms of unlabeled data settings, related models (i.e., DeepAligned, GCD, and DCCL ) still perform poorly on GRD compared to open relation extraction models (i.e., SelfORE, RoCORE, CaPL, TABs, and MatchPrompt). This is because open relation extraction models are designed specifically for relation characteristics, e.g., RoCORE constructs a relation-oriented clustering space, and TABs introduce type abstraction for relations. In our approach, we innovatively design different debiased views and the tri-view dual-space learning mechanism for relation learning, which also achieves new SOTA results.
	
	\item We find that ChatGPT underperforms in this task. It might be that ChatGPT will obtain multiple relational phrases for one relation due to various sentence expressions. Although we adopt an extra encoder to align these phrases, semantic information is lost compared to learning relations directly from the original sentences.
	
\end{enumerate}

{\bf Semantic Prediction Results on Novel Relations.} The evaluation of semantic prediction is to compare the similarity of the relational word distribution between the model-generated and the ground-truth. Since the pre-defined relation types are known, we only focus on the semantics of the novel relations. From the above baselines, we selected TABs and MatchPrompt as comparison models for this evaluation since they also utilize features in the $[$MASK$]$ position for relation learning, so we can transform their features to obtain word distributions. As illustrated in Table~\ref{sem_ave}, two main observations are as follows:
\begin{enumerate}
\item SFGRD outperforms previous models and achieves state-of-the-art results in terms of both COS and (1-KL) under all settings. In detail, when the ratio of novel relations varies, SFGRD improves the COS and (1-KL) by 32.19\%$\sim$84.45\% and 12.75\%$\sim$25.15\%, respectively. This indicates that SFGRD is capable of learning semantic-aware representations effectively. 

\item When the ratio of novel relations increases, the GRD task becomes more challenging and our model's superiority in learning semantics becomes more apparent compared to other models. There may be two reasons for this phenomenon. First, although TABs and MatchPrompt adopt $[$MASK$]$ for relation learning, which can be converted into word distributions to obtain semantics of novel relations, they were not designed to obtain semantics. Therefore, these models sacrificed semantics for classification performance in difficult cases. Second, our SFGRD adopts a class-index space for classification learning and a cluster-semantic space for semantic refinement, which can work together to ensure model performance on category and semantics even in difficult cases.

\end{enumerate}

\subsection{Ablation Studies}
\label{Ablation}

SFGRD includes two essential modules: tri-view debiased relation representation and dual-space tri-view collaborative relation learning. In this section, we conduct ablation studies with 20\% novel relations to assess the efficacy of each module. In each experiment group, settings are identical except for the variables of interest.

{\bf Effectiveness of the Tri-view Debiased Relation Representation.} In Table~\ref{views}, we evaluate the effects of each view from two perspectives. First, we examine each view individually to assess the direct effect. Second, we measure its relative effect by removing this view. Rows (1) to (6) represent the direct and relative effects of the three views. Row (7) is the result when using all three views together. 
\begin{enumerate}
	\item  {\bf Main view:}  as shown in row (1), SFGRD achieves the highest accuracy on pre-defined relations, but it also exhibits the largest deviation, e.g., 20.89\% on FewRel, between pre-defined and novel relations. Based on row (2), we find that removing the main view reduces this deviation by 6.19\%. The results demonstrate that while the main view significantly contributes to pre-defined relations, using only the main view will lead to a serious model bias towards pre-defined relations.
	
	\item  {\bf Entity debiased view:} Comparing row (6) with row (7), we find that removing this view decreases the accuracy of novel relations by 2.22\%$/$1.31\% on FewRel$/$TACRED. Hence, entity debiased views enable the model to detect novel relations more effectively. However, just using this view causes information loss since the performance on all relations is lowest in row (3).
	
	\item  {\bf Context debiased view:} Comparing row (5) with row (7), we observe that this view also improves the accuracy of novel relations as well as narrows the deviation. However, the deviation value in row (4) is the second highest, only lower than in row (1). This indicates that although the context debiased view reduces some biases, the model still suffers from other biases, e.g., entity bias.
\end{enumerate}
As a result, our SFGRD, by integrating both the entity debiased view and the context debiased view with the main view for learning relation representation, can effectively alleviate bias and improve the model performance.
\begin{table*}[]
	\caption{Comparison of SFGRD with different views on two datasets. ``$\Delta$'' represents the difference in accuracy between pre-defined relations and novel relations. The novel relation ratio is 20\%.}
	\label{views}
	\centering
	\setlength{\tabcolsep}{2.6mm}{
		\begin{tabular}{cccc|cccc|cccc}
			\toprule
			\multicolumn{1}{l}{\multirow{2}{*}{}} & \multirow{2}{*}{\begin{tabular}[c]{@{}c@{}}Main\\ View\end{tabular}} & \multirow{2}{*}{\begin{tabular}[c]{@{}c@{}}Entity \\ Debiased View\end{tabular}} & \multirow{2}{*}{\begin{tabular}[c]{@{}c@{}}Context \\ Debiased View\end{tabular}} & \multicolumn{4}{c|}{\bf FewRel}                           & \multicolumn{4}{c}{\bf TACRED}                           \\ \cmidrule{5-12} 
			\multicolumn{1}{l}{} &   &                                                                                     &                                                                                      & Pre & Nov & All   & \multicolumn{1}{c|}{ $\Delta$ (Pre - Nov)} & Pre & Nov & All   & \multicolumn{1}{c}{ $\Delta$ (Pre - Nov)} \\ \midrule
			(1)   & \usym{1F5F8}                                                                   & \usym{2715}                                                                                 & \usym{2715}                                                                                   & \bf 86.45       & 65.56 & 74.76 & 20.89                   & \bf 89.92       & 76.96 & 83.16 & 12.96                  \\ 
			(2) & \usym{2715}                                                                  & 		\usym{1F5F8}                                                                                   & \usym{1F5F8}                                                                                  & 84.50       & 69.80 & 74.30 & 14.70                   & 86.23       & 78.85 & 82.03 & 7.38                  \\ 
			(3) & \usym{2715}                                                                  & 		\usym{1F5F8}                                                                                   & \usym{2715}                                                                                  & 82.34       & 67.43 & 72.25 & 14.91                   & 85.60       & 75.15 & 80.47 & 10.45                  \\ 
			(4) & \usym{2715}                                                                & \usym{2715}                                                                              & 		\usym{1F5F8}                                                                                    & 85.59       & 69.37 & 75.41 & 16.22                   & 87.50       & 80.33 & 84.10 & 7.17                   \\ 
			(5) & \usym{1F5F8}                                                                & 		\usym{1F5F8}                                                                             & \usym{2715}                                                                               & 84.53       & 70.86 & 75.81 & 13.67                   & 86.12       & 79.78 & 82.54 & 6.34                   \\ 
			(6) &	\usym{1F5F8}                                                              & \usym{2715}                                                                               & 		\usym{1F5F8}                                                                                & 85.31       & 71.53 & 76.34 & 13.78                   & 86.91       & 81.03 & 84.26 & 5.88                   \\ 
			\rowcolor{vgray}(7) & \usym{1F5F8}                                                                  & 		\usym{1F5F8}                                                                                   & 		\usym{1F5F8}                                                                                   & 85.39       & \bf 73.75 & \bf 76.92 & 11.64                   & 86.62       & \bf 82.34 & \bf 84.95 & 4.28                   \\ \bottomrule
	\end{tabular}}
\end{table*}
\begin{table}[t]
	\caption{Average semantic prediction results of novel relations with different ratios on two datasets. Note that a smaller KL value or larger COS value is preferable. To avoid confusion, we represent KL as (1-KL), making the two indicators consistent, i.e., higher values are always better.  }
	\label{sem_ave}
	\renewcommand\arraystretch{1}{
		\setlength{\tabcolsep}{2.1mm}{
			\begin{tabular}{c|c|cc|cc}
				\toprule
				\multirow{2}{*}{Nov/All} & \multirow{2}{*}{Model} & \multicolumn{2}{c|}{FewRel} & \multicolumn{2}{c}{TACRED} \\ \cline{3-6} 
				&                        & COS           & (1-KL)         & COS          &(1-KL)        \\ \midrule
				\multirow{4}{*}{20\%}    & TABs                   & 0.305        & 0.687        & 0.290        & 0.587       \\
				& MatchPrompt            & 0.342        & 0.703        & 0.320        & 0.604       \\
				& \bf SFGRD                  & \bf0.479        & \bf0.793        & \bf 0.423        & \bf0.681      \\
				&  \multirow{2}{*}{\bf$\Delta$}                   & $\uparrow$0.137       & $\uparrow$0.090        & $\uparrow$0.103      & $\uparrow$0.077       \\
				&               &\cellcolor{vgray}        40.05\%              & \cellcolor{vgray}   12.8\%             &\cellcolor{vgray}   32.19\%              &\cellcolor{vgray}   12.75\%             \\ \midrule
				\multirow{4}{*}{50\%}    & TABs                   & 0.235        & 0.630        & 0.191        & 0.510       \\
				& MatchPrompt            & 0.277        & 0.652        & 0.239        & 0.560       \\
				& \bf SFGRD                  & \bf0.428        & \bf 0.766        & \bf 0.405        & \bf 0.653       \\
				& \multirow{2}{*}{\bf$\Delta$}                   & $\uparrow$0.151       & $\uparrow$0.114       & $\uparrow$0.166      & $\uparrow$0.093         \\
				&                       &\cellcolor{vgray}   54.51\%              & \cellcolor{vgray}   17.48\%             &\cellcolor{vgray}   69.46\%              &\cellcolor{vgray}   16.60\%             \\ \midrule
				\multirow{5}{*}{80\%}    & TABs                   & 0.184        & 0.595       & 0.203        & 0.470       \\
				& MatchPromt             & 0.219        & 0.628        & 0.241        & 0.493       \\
				& \bf SFGRD                  & {\bf 0.404}        & \bf 0.749        & \bf 0.413        & \bf 0.617      \\ 
				& \multirow{2}{*}{\bf$\Delta$}                   & $\uparrow$0.185       & $\uparrow$0.121        & $\uparrow$0.172       & $\uparrow$0.124         \\
				&                       &\cellcolor{vgray}   84.45\%              & \cellcolor{vgray}   19.26\%             &\cellcolor{vgray}   71.37\%              &\cellcolor{vgray}   25.15\%             \\\bottomrule
				
	\end{tabular}}}
\end{table}

\begin{table}[]
	\caption{Accuracy (\%) on different variants of SFGRD on two datasets. ``$\Delta$'' represents the change value of the variants compared with SFGRD. The novel relation ratio is 20\%.}
	\label{model design} 
	\setlength{\tabcolsep}{2.1mm}{
		\begin{tabular}{c|ccc|ccc}
			\toprule
			& \multicolumn{3}{c|}{FewRel} & \multicolumn{3}{c}{TACRED}  \\ \cmidrule{2-7} 
			& Pre & Nov & All   & Pre & Nov & All   \\ \midrule
			\bf SFGRD    & 85.39       & \bf 73.75 & \bf 76.92 & 86.62       & \bf 82.34 & \bf 84.95 \\ \midrule
			SFGRE$^{-Contra}$ & \bf 86.92       & 71.12 & 74.22 & \bf 89.13       & 77.28 & 81.57 \\
			$\Delta$       & \cellcolor{vgray}$\uparrow$1.53      & \cellcolor{vgray}$\downarrow$2.63 & \cellcolor{vgray}$\downarrow$2.7  &  \cellcolor{vgray}$\uparrow$2.51       & \cellcolor{vgray}$\downarrow$5.06 & \cellcolor{vgray}$\downarrow$3.38 \\
			SFGRD$^{-Align}$ & 85.15       & 70.67 & 73.36 & 86.41       & 78.62 & 80.70 \\
			$\Delta$       &  \cellcolor{vgray}$\downarrow$0.24       & \cellcolor{vgray}$\downarrow$3.08 & \cellcolor{vgray}$\downarrow$3.56 &  \cellcolor{vgray}$\downarrow$0.21       & \cellcolor{vgray}$\downarrow$3.72 & \cellcolor{vgray}$\downarrow$4.25 \\
			SFGRD$^{-Select}$ & 85.25       & 70.73 & 73.17 & 84.77       & 77.41 & 78.59 \\
			$\Delta$        & \cellcolor{vgray}$\downarrow$0.14       & \cellcolor{vgray}$\downarrow$3.02 & \cellcolor{vgray}$\downarrow$3.75 & \cellcolor{vgray}$\downarrow$1.85       &\cellcolor{vgray} $\downarrow$4.93 & \cellcolor{vgray}$\downarrow$6.36 \\ \bottomrule
	\end{tabular}}
\end{table}

{\bf Effectiveness of the Dual-space Tri-view Collaborative Relation Learning.} We construct three variants of SFGRD to verify the effectiveness of different components in this module. 1) SFGRD$^{-Contra}$, the variant model that remove the contrastive learning across three views. We achieve this by eliminating loss functions ${\ell_s}$ and ${\ell_p}$ in eqs. (\ref{e2}) and (\ref{consis}). 2) SFGRD$^{-Align}$, the variant model that remove the alignment strategy. We achieve this by just aligning the clustering results of tri-view in the cluster-semantic space without using the anchor labels from class-index space. 3) SFGRD$^{-Select}$, the variant model that remove the selection strategy. We achieve this by eliminating the eq. (\ref{sele}). The results of SFGRD and its three variants on two datasets are shown in Table.~\ref{model design}, we can make the following observations.
\begin{enumerate}

\item When removing contrastive learning across three views, the performance on pre-defined relations increases by 1.53\%/2.51\% on FewRel/TACRED, but on novel relations, it decreases by 2.63\%/5.06\%. This indicates that contrastive learning alleviates the bias against pre-defined relations and allows the model to learn effective representations, especially for novel relations.

\item Using anchor labels in the class-index space to align the clustering results of three views is more effective than relying solely on self-alignment. For example, after removing the alignment strategy, the model performance on novel relations dropped by 3.08\% on the FewRel dataset. This confirms the effectiveness of class-index space in learning index information. 

\item The selection strategy allows the model to fully consider the differences between clustering results from tri-view, select reliable pseudo-labels for further training.

\item  When either the alignment strategy or the selection strategy is removed, performance degradation on novel relations is significantly greater than pre-defined relations. This may be because pre-defined relations can also be learned from annotated data, which increases their robustness. Novel relations, however, learn entirely from unlabeled data, so they suffer significant performance losses upon removing alignment and selection strategies. This also illustrates the importance of these strategies in improving the robustness of the model for handling unlabeled data.
\end{enumerate}

\begin{figure*}[t]
	\centering
	\begin{subfigure}{0.3\textwidth}
		\centering
		\includegraphics[width=\linewidth]{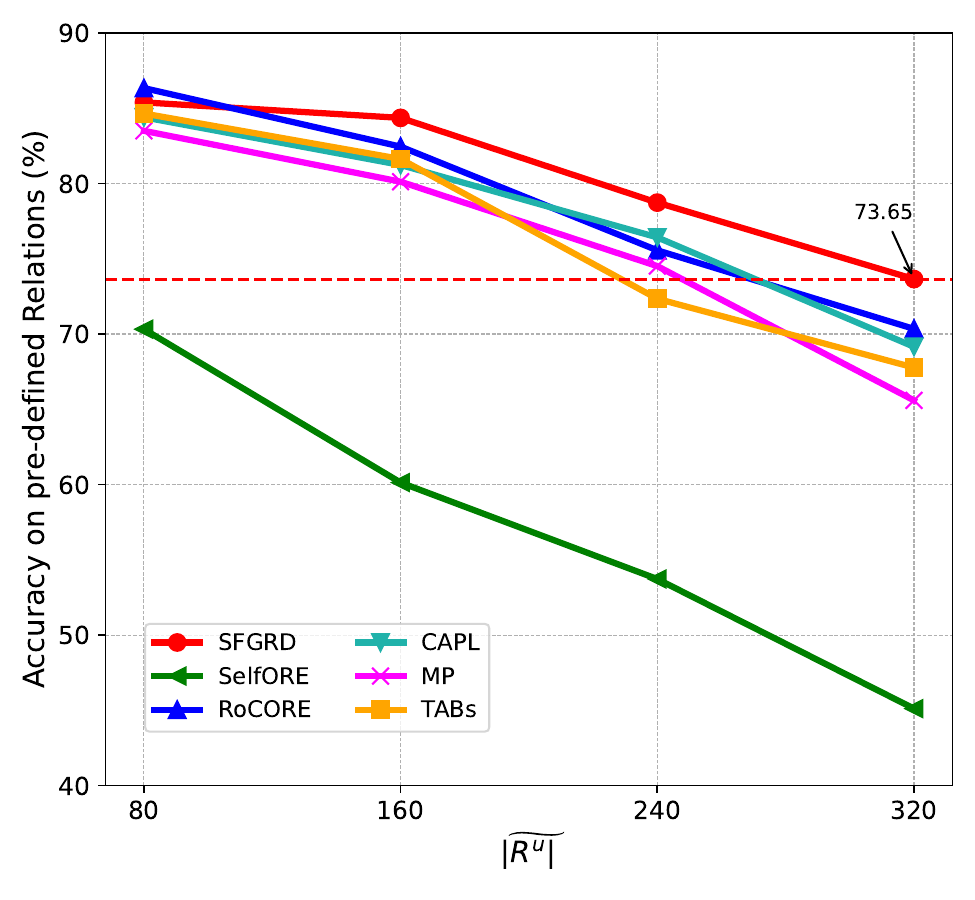}
		\caption{Pre-defined}
	\end{subfigure}%
	\hfill
	\begin{subfigure}{0.3\textwidth}
		\centering
		\includegraphics[width=\linewidth]{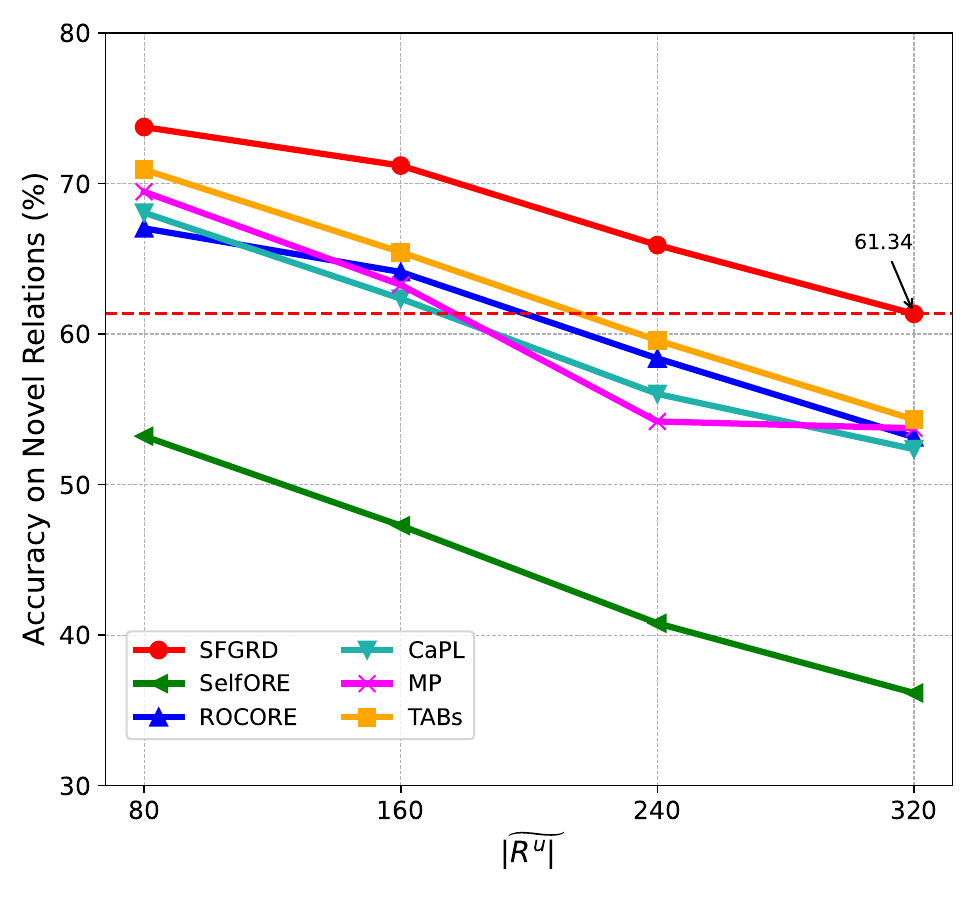}
		\caption{Novel}
	\end{subfigure}%
	\hfill
	\begin{subfigure}{0.3\textwidth}
		\centering
		\includegraphics[width=\linewidth]{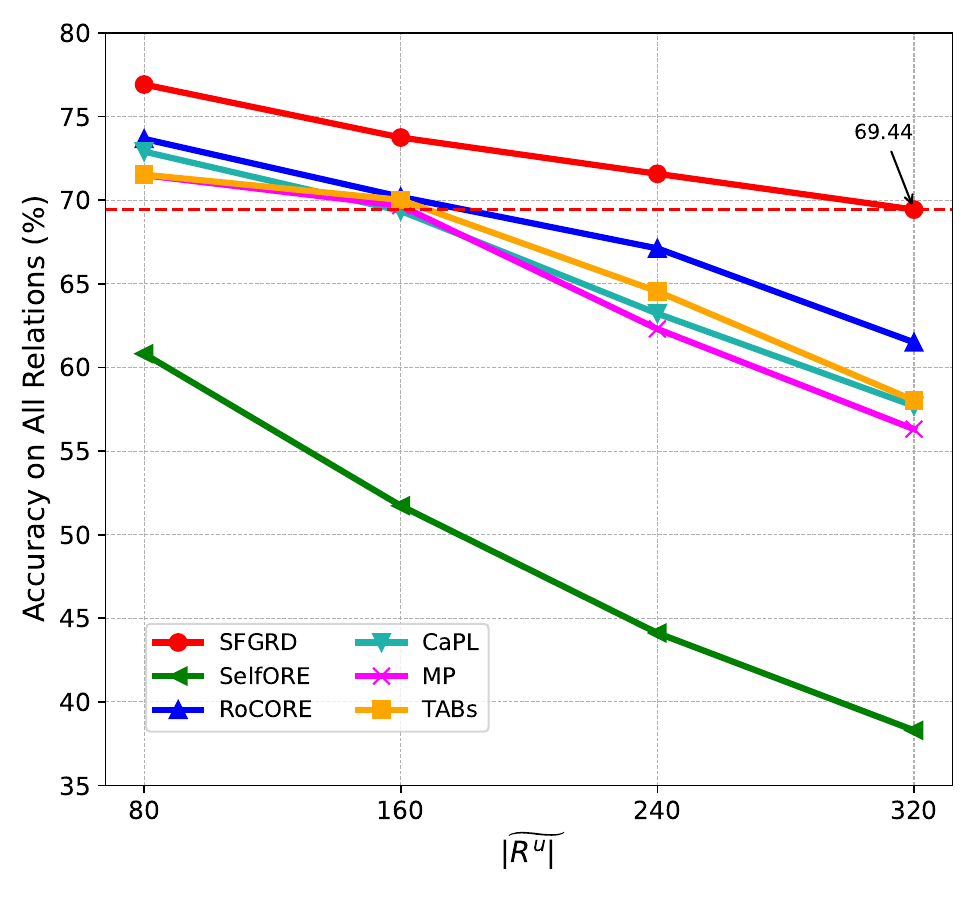}
		\caption{All}
	\end{subfigure}
	\caption{Comparison of different models accuracy against different initiation numbers of clusters on FewRel dataset. The novel relation ratio is 20\%.}
	\label{Est_K}
\end{figure*}

\subsection{Estimating the Relation Number in Unlabeled data}
\label{cluster_num}
As the FewRel dataset has more relations than TACRED, so we use it as an example to validate the performance of SFGRD when the number of relations in the unlabeled dataset is unknown. We adopt the easy and effective algorithm introduced in DeepAligned~\cite{DBLP:conf/aaai/Zhang0LL21} to tackle the problem of estimating the number of relations $|R^u|$ in unlabeled data. To achieve this, we set a large value $\widetilde{|R^u|}$ as the initiation number of relations, and apply K-means with $\widetilde{|R^u|}$ cluster centers on unlabeled data. Suppose that the real clusters tend to be dense even with $\widetilde{|R^u|}$, and the instance number of more confident clusters is larger than the expected cluster average size, i.e., $N_2/\widetilde{|R^u|}$. On this basis, we drop the low confident clusters and obtain the estimated number of relations. We take FewRel dataset as an example and vary $\widetilde{|R^u|}$ from the ground-truth number to four times of it, i.e., $\widetilde{|R^u|}=\{80,160,240,320\}$, and select five open relation extraction methods for comparison. The estimated number of relations and the accuracy of different models are reported in Table~\ref{Est} and Fig.~\ref{Est_K}, respectively.

SFGRD achieves lower error rates than other models. This indicates that our model can learn effective representations, which is beneficial for estimating the number of relations. As redundant classes may result in many fine-grained clusters for one original relation, the performance of models decreases with the value of $\widetilde{|R^u|}$ increases. Despite this, we can see that the accuracy of our SFGRD declines slower compared to the others. There may be two reasons.  {\bf First}, as mentioned above, our SFGRD benefits from a more accurate estimated cluster number. {\bf Second}, the SFGRD framework, combined with its alignment and selection strategies, stabilizes its performance even when the estimated cluster number is less accurate. For example, when $\widetilde{|R^u|}=320$, the number of relations predicted by SFGRD is 97. This value is close to the number of relations predicted by other models at $\widetilde{|R^u|}=240$. Therefore, we plot red dotted lines as shown in Figure~\ref{Est_K}, and find that the accuracy of SFGRD at $\widetilde{|R^u|}=320$ outperforms other models on both novel and all relations at $\widetilde{|R^u|}=240$. 
\begin{table}[t]
	\caption{Estimation of the number of relations on FewRel dataset. The ground-truth number of relations is $|R^u|=80$ and the novel relation ratio is 20\%. ``Error'' denotes the error rate when comparing predicted numbers with the ground-truth numbers.}
	\label{Est}
	
	\setlength{\tabcolsep}{0.62mm}{
		\begin{tabular}{c|cc|cc|cc}
			\toprule
			\multirow{2}{*}{} & \multicolumn{2}{c|}{$\widetilde{|R^u|}$=160} & \multicolumn{2}{c|}{$\widetilde{|R^u|}$=240} & \multicolumn{2}{c}{$\widetilde{|R^u|}$=320} \\ \cmidrule{2-7} 
			& $|R^u|$ (Pred)      & Error       & $|R^u|$ (Pred)      & Error      & $|R^u|$ (Pred)      & Error   \\ \midrule
			\bf SFGRD                            & \bf 74           & \bf 7.5\%       & \bf 91           & \bf 13.8\%      & \bf 97           & \bf 21.3\%     \\\midrule
			SelfORE                          & 90           & 12.5        & 104          & 30.0\%      & 116          & 45.0\%     \\
			RoCORE                           & 72           & 10.0\%      & 95           & 18.8\%      & 105          & 31.3\%     \\
			CaPL                             & 70           & 12.5\%      & 97           & 21.3\%      & 108          & 35.0\%     \\
			MatchPrompt                      & 70           & 12.5\%      & 98           & 22.5\%      & 110          & 37.5\%     \\
			TABs                             & 73           & 8.8\%       & 94           & 17.5\%      & 103          & 28.8\%     \\ \bottomrule
	\end{tabular}}
\end{table}
\begin{table*}[t]
	\caption{\label{diss}Top predicted relational words for some novel relations on FewRel. For each relation, we count the frequency of words appearing in the top 3 among 100 instances in the {\ttfamily[MASK]} position. Note that a high ACC and high COS indicates that the current label index accuracy and semantic quality cosine similarity are higher than the corresponding median values, respectively, and vice versa.}
	\small
	\centering
	\setlength{\tabcolsep}{2.2mm}{
	\begin{tabular}{cc|c|l|l}
		\toprule
		\bf ACC                  & \bf COS                  & \bf Relations     & {\bf Examples} ($ < S_1 >$. {\underline{head entity}}{\ttfamily [MASK]} \uwave {tail entity}. )                                                      & \bf Top Predicted Relational Words \\ \midrule
		\multirow{2}{*}{High} & \multirow{2}{*}{High} & crosses       & \underline{Sir Leo Hielscher Bridges} cross over the  \uwave{Brisbane River}.            & cross, crosses, crossing         \\\cmidrule{3-5}
		&                       & constellation & \underline{NGC 451} is a spiral galaxy located in the constellation \uwave {Pisces}.    & constellation, borders, edges    \\ \midrule
		\multirow{2}{*}{High} & \multirow{2}{*}{Low}  & voice type    & \underline{Carl Tanner} is an American operatic \uwave {tenor}.                         & \#\#ored, \#\#-1, \#\#nier       \\ \cmidrule{3-5} 
		&                       & military rank & \underline{Francis Pollet} is \uwave {General officer}.                                 & \#\#ibar, \#\#anga, \#\#oliver   \\ \midrule
		\multirow{2}{*}{Low}  & \multirow{2}{*}{High} & Spouse        & \underline{Catherine} is marred to \uwave {Duke Henry}.                                 & marriages, spouse, married       \\ \cmidrule{3-5} 
		&                       & Mother        & \uwave {Marsy}'mother, \underline{Marcella}, suffered a heart attack at the parole.   & son, mother, birth               \\ \midrule
		\multirow{2}{*}{Low}  & \multirow{2}{*}{Low}  & member of     & \underline{Peter Trudgill}, \uwave {FBA}, is a sociolinguist, academic, and author.     & is, \#\#lle, \#\#osity           \\ \cmidrule{3-5} 
		&                       & part of       & Strutt lives on the island of \underline{South Ronaldsay} in \uwave {Orkney}, Scotland. & in, \#\#under, \#\#nary          \\ \bottomrule
	\end{tabular}}
\end{table*}


\subsection{Further Analysis }
\label{case}
\begin{figure}
	\centering
	\includegraphics[width=0.8\linewidth]{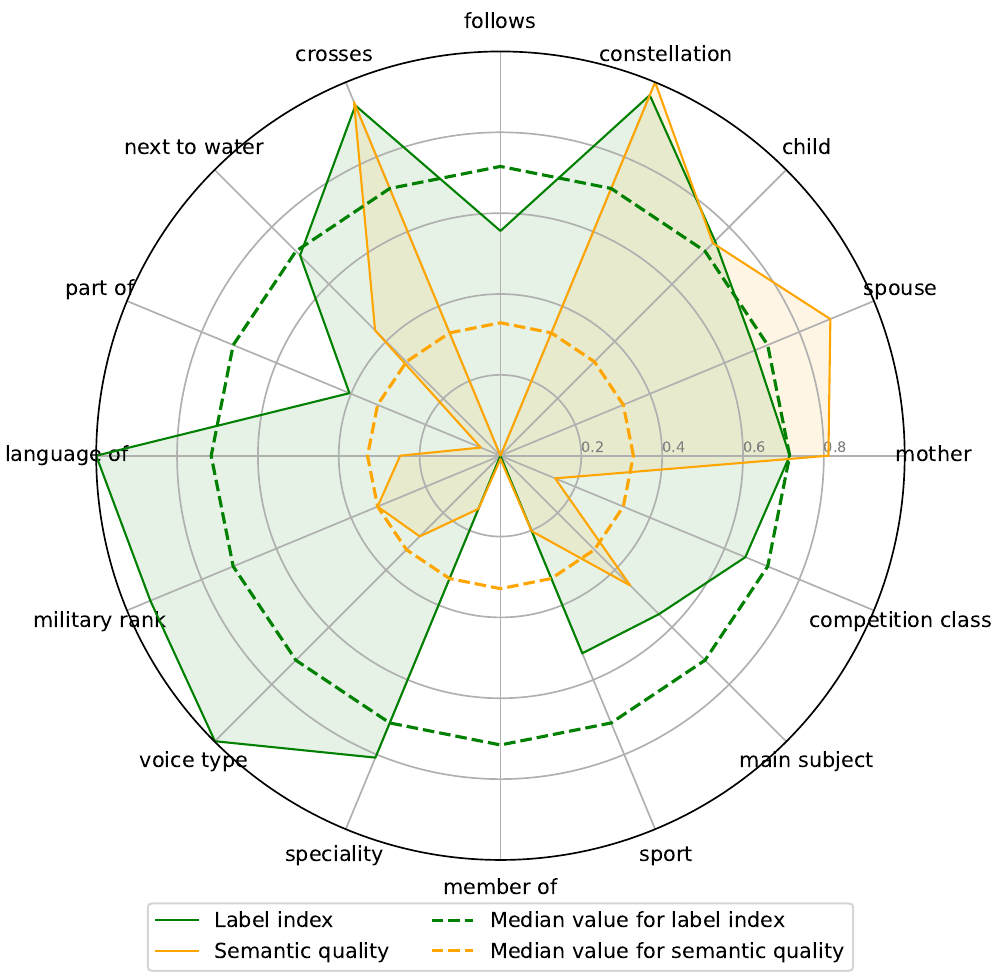} 
	\caption{Normalized label index accuracy and semantic cosine similarity of different novel relations on FewRel. The dashed lines represent the median values.}
	\label{radr}
\end{figure}
In this section, we explore the performance of label index accuracy and semantic quality on different relations. Specifically, we examine 16 novel relationships within the FewRel dataset. To effectively illustrate the performance distinctions of each relation on a unified comparative scale, we have normalized the accuracy (ACC) and cosine similarity (COS) values respectively for these relations. The results are visually represented in a radar chart, as shown in Fig~\ref{radr}.

{\bf Label Index Accuracy of Different Relations.}
The performance of the label index is related to two factors. First, the prominent features play a crucial role. Relations like ``Constellation'', `` Language of'', ``voice type'', and ``Military rank'' are both salience and have high accuracy. In contrast, ``member of'' or ``Part of'' has low performance due to their wide feature scope, e.g., part of an island, a literary, or a sporting season. Second, the degree of similarity to other relations impacts accuracy. We notice that relations like ``Mother'', ``Child'' and ``Spouse'' have a relatively moderate accuracy. This may be due to the similarity of sentences across these relations, where altering the positions of the head and tail entities in the same sentence can lead to different relations.

{\bf Semantic Quality of Different Relations}
To better evaluate the semantic quality and compare the correlation between semantic quality and label index accuracy, we use the median values as the boundaries (the dashed lines in Fig.~\ref{radr}) to select some relations based on different label index accuracy and semantic cosine similarity combinations. We present the top predicted relational words of these relations in Table~\ref{diss}. The detailed analysis is as follows.
\begin{enumerate}
	\item  We find that higher semantic cosine similarity demonstrates the predicted words are closer to the ground-truth relations. For example, the top predicted words ``cross'', ``crosses'' and `` crossing'' are almost identical in meaning to  the true relation ``Crosses''.
	\item  The high accuracy of the label index does not necessarily equate to high semantic quality. For example, relations such as ``voice type'' and ``Military rank'' have high label index accuracy, but their semantic quality may not be equally high. This phenomenon may be attributed to the fact that the semantic quality mainly depends on the presence of explicit words that are directly associated with the relation within sentences. In other words, Even though some head or tail entities in the sentence (like 'NGC451') might be difficult to comprehend, the appearance of key terms such as ``constellation'' can help the model predict the correct relational words. In contrast, as we mentioned above, label index accuracy relies more on whether the features of the relation are sufficiently prominent.
	
	\item  For relational sentences lacking both prominent inherent features and keywords of relations, it is natural to observe both lower label index accuracy and semantic quality.
	
\end{enumerate}


\section{Conclusion}
In this article, we introduce the GRD task that identifies open-world unlabeled instances into a pre-defined relation or discovers novel relations by not only assigning instances to clusters but also providing specific meanings for these clusters.  We propose a novel framework, SFGRD, which learns from semi-factual and employs a two-stage process including a tri-view debiased relation representation module and a dual-space tri-view collaborative relation learning module for this task. Extensive experimental results on two datasets show that SFGRD surpasses the state-of-the-art models not only in accuracy for relation class learning but also in the quality of relation semantic learning. In the future, we will continue to improve the accuracy and quality of extraction results in the open world.

\section*{Acknowledgments}
This work was supported by National Key Research and
Development Program of China (2022YFC3303600), National
Natural Science Foundation of China (62137002, 62192781,
62293553, 61937001, 62250066, and 62106190), Innovative
Research Group of the National Natural Science Foundation
of China (61721002), ``LENOVO-XJTU'' Intelligent Industry
Joint Laboratory Project, Foundation of Key National Defense
Science and Technology Laboratory (6142101210201), Natural
Science Basic Research Program of Shaanxi (2023-JCYB-
593), the Youth Innovation Team of Shaanxi Universities,
the Fundamental Research Funds for the Central Universities
(xhj032021013-02), XJTU Teaching Reform Research Project
``Acquisition Learning Based on Knowledge Forest''.

\bibliographystyle{IEEEtran}
\bibliography{bare_jrnl_new_sample4}
\newpage

\vfill

\end{document}